\documentclass[letterpaper, 10 pt, conference]{ieeeconf}  

\IEEEoverridecommandlockouts                              

\usepackage{cite}
\usepackage{amsmath,amssymb,amsfonts}
\usepackage{algorithmic}
\usepackage{graphicx}                                                          
\usepackage{float} 
\usepackage{subfig}
\usepackage{textcomp}
\usepackage[table,xcdraw]{xcolor}
\usepackage{booktabs}
\usepackage{multirow}
\usepackage{multicol}
\usepackage{threeparttable}
\usepackage{lipsum}
\usepackage{color}
\usepackage[linesnumbered, ruled]{algorithm2e}
\makeatletter
\let\NAT@parse\undefined
\makeatother
\usepackage[pagebackref, colorlinks]{hyperref}
\SetKwRepeat{Do}{do}{while}%
\SetKwRepeat{TimeUpdate}{Time update}{end}%
\SetKwRepeat{MeasurementUpdate}{Measurement update}{end}%
\SetKwRepeat{ConstraintsUpdate}{Constraints update}{end}%

\title{\LARGE \bf
RELEAD: Resilient Localization with Enhanced LiDAR Odometry in Adverse Environments
}

\author{Zhiqiang Chen, Hongbo Chen, Yuhua Qi$^{\star}$, Shipeng Zhong, Dapeng Feng, \\Jin Wu, Weisong Wen and Ming Liu
\thanks{This work was supported by the project “Development of sensing data collection and experimental platform for unmanned swarm system” from AMOVLAB. Corresponding Author: Yuhua Qi.}
\thanks{Zhiqiang Chen, Hongbo Chen, Yuhua Qi, Shipeng Zhong and Dapeng Feng are with the School of Systems Science and Engineering, Sun Yat-sen University, Guangzhou 510006, China (e-mail: \{chenzhq56, zhongshp5, fengdp5\}@mail2.sysu.edu.cn, \{chenhongbo, qiyh8\}@mail.sysu.edu.cn)}
\thanks{J. Wu and M. Liu are with Department of Electronic and Computer Engineering, Hong Kong University of Science and Technology, Hong Kong (e-mail: jin\_wu\_uestc@hotmail.com, eelium@ust.hk).}
\thanks{W. Wen is with Department of Aeronautical and Aviation Engineering, Hong Kong Polytechnic University, Hong Kong (e-mail: Welson.wen@polyu.edu.hk).}
}

\begin{document}

\maketitle
\thispagestyle{empty}
\pagestyle{empty}

\begin{abstract}
        LiDAR-based localization is valuable for applications like mining surveys and underground facility maintenance. However, existing methods can struggle when dealing with uninformative geometric structures in challenging scenarios. This paper presents RELEAD, a LiDAR-centric solution designed to address scan-matching degradation. Our method enables degeneracy-free point cloud registration by solving constrained ESIKF updates in the front end and incorporates multisensor constraints, even when dealing with outlier measurements, through graph optimization based on Graduated Non-Convexity (GNC). Additionally, we propose a robust Incremental Fixed Lag Smoother (rIFL) for efficient GNC-based optimization. RELEAD has undergone extensive evaluation in degenerate scenarios and has outperformed existing state-of-the-art LiDAR-Inertial odometry and LiDAR-Visual-Inertial odometry methods.
\end{abstract}

                                                                        
%
\section{INTRODUCTION}
\subsection{Motivation}
Localization is a fundamental requirement in numerous robotic applications, including search-and-rescue missions, mining surveys, and maintenance of underground facilities. These applications demand robust positioning in GNSS-denied environments for effective navigation. LiDAR-based odometry, leveraging accurate point clouds \cite{Zhang2014LOAMLO,Behley2018EfficientSS,liu2021balm, li2021towards}, generally provides reliable state estimation via point cloud registration. However, these methods encounter challenges in perceptually complex settings, such as geometrically uninformative environments like tunnels and open fields. 
The absence of distinctive LiDAR features in such symmetrical and self-similar environments hampers registration performance and leads to optimization divergence along inadequately constrained directions \cite{Tuna2022XICPLL}, commonly referred to as degeneracy.

\subsection{Challenges}

\begin{figure}[t]
        \centering
        \includegraphics[width=0.47\textwidth]{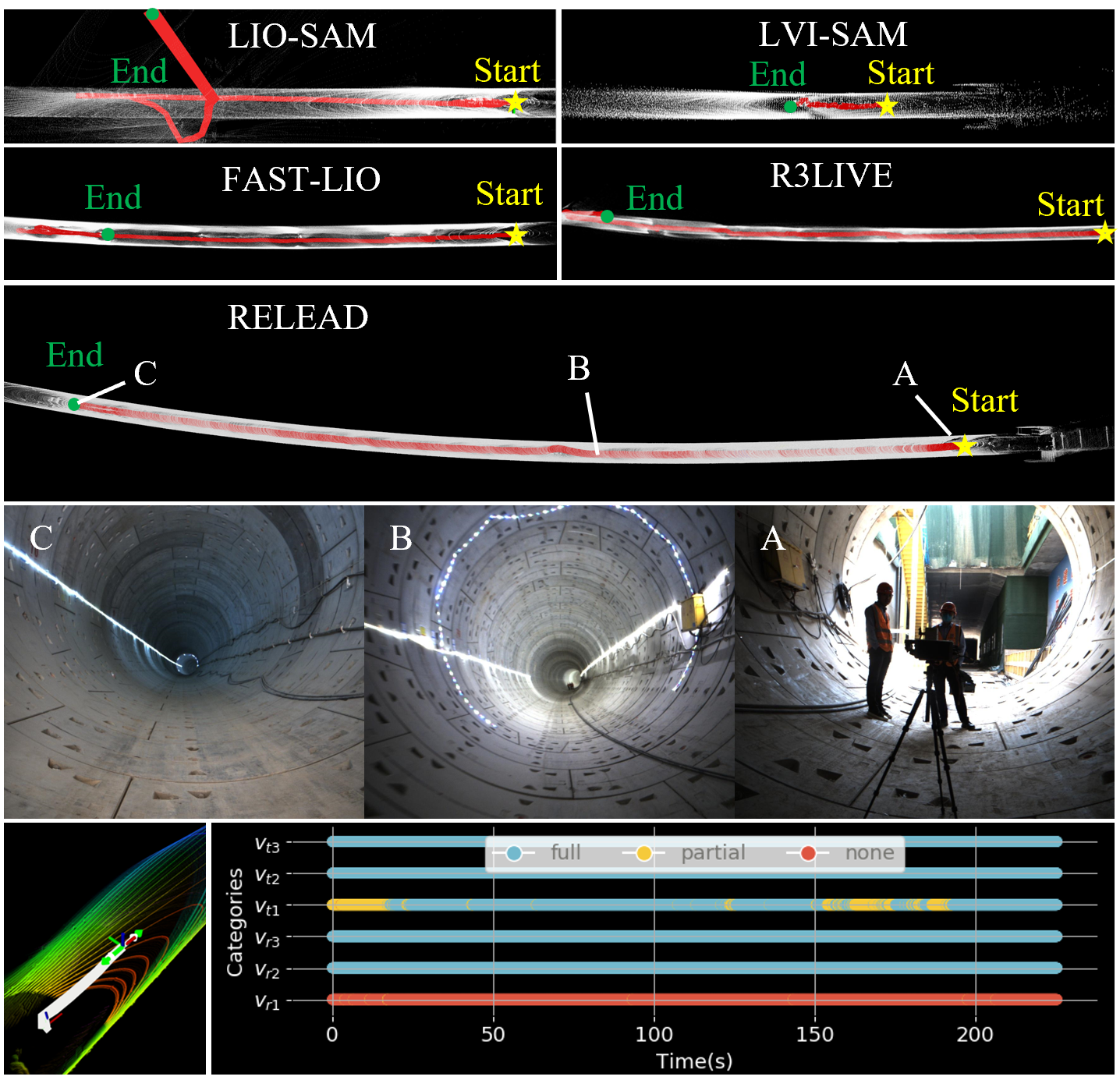}
        \caption{
        Comparison of methods for metro tunnel reconstruction and associated trajectories (red lines). In this degenerate scenario, other methods fail due to the lack of distinctive features, resulting in a slipping trajectory, while RELEAD successfully reconstructs the scene. In the bottom row: the left panel illustrates degradation detection within the tunnel, with arrows indicating degradation direction. The right panel displays estimated localizability categories for each direction.
        }
        \label{fig:shenzhentunnel}
        \vspace{-0.7cm} 
\end{figure}

The crux of the LiDAR SLAM degradation problem lies in the fact that geometric constraints in the degenerate direction are almost indistinguishable from noise. Hence, addressing degeneracy typically involves two approaches: i) actively managing degenerate situations through detection and mitigation and ii) introducing new constraints through additional sensing modalities. Despite promising results in previous work on degeneracy-related localization, significant challenges persist.

\subsubsection{Robust Odometry against Degradation}
Most existing approaches for detecting degeneracy, exemplified by \cite{Zhang2016OnDO,ebadi2021dare, Hinduja2019DegeneracyAwareFW,ding2021degeneration}, primarily rely on pose optimization characteristics to identify degeneracy and compare metrics against specific thresholds. These methods often incorporate solution remapping or constrained optimization techniques to mitigate degeneracy. However, most of these methods require manual threshold adjustments in different applications.
Moreover, using eigenvalues in these approaches may demonstrate incongruous outcomes across various settings and sensors, impeding generalizability.
Geometric methods X-ICP \cite{Tuna2022XICPLL} achieves environment-independent detection in the presence of priori pose; it gauges the sensitivity of point and surface-normal pairs to optimization states through constraint strength. However, the method targets specific modules within the SLAM pipeline and does not consider using additional sensor information to augment the accuracy of the initial guess.

\subsubsection{Multi-Sensor Fusion Methods}
To deal with the degeneration caused by the incomplete constraints of LiDAR sensors, methods such as \cite{Grter2018LIMOLV, Zhu2020CamVoxAL, Shan2021LVISAMTL,lin2022r3, nubert2022graph, zhao2021super, khedekar2022mimosa, wang2022rail, Khattak2020ComplementaryMS, Palieri2020LOCUSAM,reinke2022locus} primarily focus on different fusion paradigms to leverage multi-sensor data for improved accuracy and robustness. A typical categorization based on the mutual relations of observations from different modalities is discussed below.
Super-observation pipeline like CamVox\cite{Zhu2020CamVoxAL} aims at synchronized points and images to form RGB-D frames, so all particular modalities observations always belong to one single state realization.
Conditionally independent pipeline considers both synchronized and asynchronous multimodal measurements, and the motion model establishes the cross-modality link; the update step is separable in terms of modality. 
R3LIVE\cite{lin2022r3} and LVI-SAM\cite{Shan2021LVISAMTL} both get pose estimation via IMU preintegration, visual-inertial odometry, and LiDAR-inertial odometry separately, then fuse the three types of estimation either in a filter or a pose graph.
V-LOAM\cite{Zhang2018LaservisualinertialOA} utilizes a sequential processing pipeline where motion estimation begins with IMU prediction and incorporates a visual-inertial estimator loosely coupled with LiDAR depth association.
The HeRO algorithm, a parallel system \cite{SantamariaNavarro2020TowardsRA}, improves estimation accuracy by smoothly transitioning between different sources of odometry.
Even though robust multi-sensor fusion systems above are designed to withstand degraded sensing, they require laborious manual parameter tuning and exhibit limited resilience to outlier measurements and environmental changes \cite{ebadi2022present}.

\subsection{Contributions}
To enhance the accuracy and resilience of SLAM methods across diverse environments and improve real-world robustness, we introduce a LiDAR-centric localization and mapping approach. This approach integrates an environment-agnostic degeneracy detection and mitigation module with a novel failure-tolerant multi-sensor fusion framework. The hierarchical combination of the filter and factor graph framework ensures efficient utilization of multi-sensor information for accurate initial pose estimation and additional constraints on the system state, thereby enhancing overall performance.
Specifically, our contributions are as follows:
\begin{enumerate}
        \item We propose a degeneracy-eliminated LiDAR-Inertial odometry. Specifically, the alignment strength for 6 DOF is analyzed utilizing the degradation detection module. The Constrained Error-State Iterated Extended Kalman Filter (CESIKF) can offer drift-free pose updates by combining the detection of ill-conditioned directions.
        \item We deliver a sensor integration module to keep the system state well-constrained in LiDAR-degenerate scenarios. The factor graph optimization is designed to handle delayed sensor measurements, and the graduated non-convexity factor (GNC factor) guarantees outlier-free sensor fusion.
        \item To efficiently solve the GNC optimization problem, we propose a robust Incremental Fixed Lag Smoother combining the sliding window factor graph framework and GNC factor. This approach enables robust multi-modal sensor fusion while ensuring online efficiency.
        \item Experiments are conducted on both open datasets and our private device data. Results show that RELEAD can handle challenging sensor-degenerated environments and provide more accurate pose estimation.
\end{enumerate}

\section{METHOD}
\begin{figure}[t]
        \centering
        \includegraphics[width=0.44\textwidth]{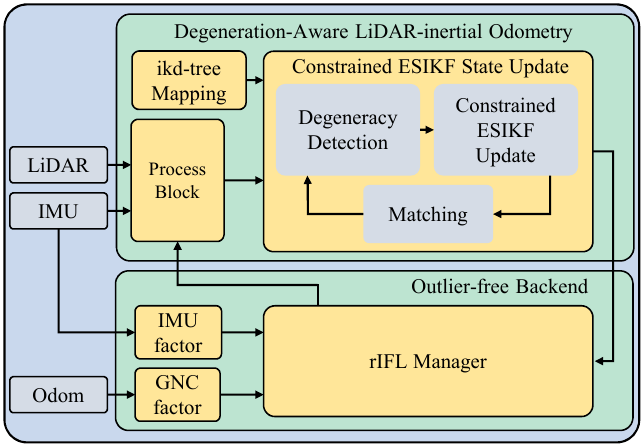}
        \caption{System overview of RELEAD consists of three parts: process block, constrained ESIKF update process and graph-based sensor integration module.}
        \label{fig:pipeline}
        \vspace{-0.7cm} 
\end{figure}

RELEAD introduces degeneration-aware LiDAR-inertial odometry with an outlier-free sensor integration module, ensuring dependable fusion of supplemental odometry. The system's architecture, depicted in Fig.\ref{fig:pipeline}, comprises three main components:
i) \emph{Process Block}: This component is responsible for acquiring the prior distribution and undistorting point clouds. It propagates IMU data using a discrete kinematic model and applies B-spline interpolation to correct individual point distortions for improved motion compensation. The propagation result is then updated with pose measurements from rIFL to provide a reliable prior for state estimation.
ii) \emph{CESIKF-based State Estimation Module}: This module generates drift-free pose updates using three key components: a degeneracy detection module, scan-matching residual calculation, and constrained state update. Based on the scan-matching result, the degeneracy detection module identifies degenerate directions and computes additional constraints if degeneracy is detected. State estimation is solved using a constrained ESIKF approach incorporating prior distribution, measurement residuals, and additional constraints.
iii) \emph{Graph-based Multi-Sensor Fusion Backend}: This backend ensures the system state is
well-constrained and provides more precise estimations through multiple modality constraints. It employs an IMU-centric sensor fusion architecture, including mechanisms for handling outliers and flexibility to fuse delayed measurements using robust Incremental Fixed Lag Smoother.

\subsubsection*{Notation} Assuming the known extrinsic ${ }^I \mathbf{T}_L$ relationship between LiDAR and IMU, we denote the IMU frame as $I$ 
and consider it the body frame, with the initial body frame designated as the global frame $G$.
We define the state vector $\mathbf{x}$ as:
\begin{equation}
        \small
        \begin{aligned}
                & \mathbf{x} \triangleq\left[\begin{array}{llllll}
                        { }^G \mathbf{R}_I^T & { }^G \mathbf{t}_I^T & { }^G \mathbf{v}^T & \mathbf{b}_{\mathbf{g}}^T & \mathbf{b}_{\mathbf{a}}^T & { }^G \mathbf{g}^T
                \end{array}\right]^T, 
        \end{aligned}
\end{equation}
where ${ }^G \mathbf{R}_I$, ${ }^G \mathbf{t}_I$ and ${ }^G \mathbf{v}_I$ denote the IMU attitude, position, and velocity in the global frame. $\mathbf{b}_{\mathbf{a}}$  and $\mathbf{b}_{\mathbf{g}}$ represent the gyroscope and accelerometer biases. ${ }^G \mathbf{g}$ corresponds to the gravity vector in the global frame.

\subsection{Process Block}
The process block includes two main compensates: pointcloud undistortion and prediction model.
\subsubsection{Pointcloud Undistortion}
Accumulating LiDAR point clouds with continuous LiDAR motion leads to non-rigid distortion, adversely impacting localization accuracy and degeneracy detection. To rectify this distortion, RELEAD estimates the transform of each LiDAR point relative to the last point in the same frame using propagation and B-spline interpolation. B-spline interpolation overcomes the limitations associated with IMU frequency and avoids approximate linear interpolation.

Here, we use the forward propagation as prior to predicting the state $\hat{\mathbf{x}}_{i+1}$ and its covariance $\hat{\boldsymbol{\Sigma}}_{\delta \hat{\mathbf{x}}_{i+1}}$, at each IMU input $\mathbf{u}_i$, where $i$ represents the index of IMU measurements. Specifically, we propagate the state in the same way as \cite{Xu2020FASTLIOAF}:
\begin{equation}
        \widehat{\mathbf{x}}_{i+1}=\widehat{\mathbf{x}}_i \boxplus\left(\Delta t \mathbf{f}\left(\widehat{\mathbf{x}}_i, \mathbf{u}_i, 0\right)\right) ,
\label{equation:forward_propagation}
\end{equation}
where $\Delta t$ represents the IMU sampling interval, $\mathbf{u}$ is the IMU measurement, and $\boxplus$ denotes the encapsulated "boxplus" operation on the manifold \cite{hertzberg2013integrating}.
Building on this propagation, we employ B-spline interpolation \cite{jung2023asynchronous} to correct in-scan motion.

\subsubsection{Prediction Model}
The propagation is affected by IMU measurement noises and bias estimation errors. In order to enhance the accuracy of scan-matching and degeneracy detection, we consider the optimized pose obtained from the pose graph (Section \ref{subsection:backend}) as a global pose measurement. Consequently, the final prior distribution combines both the IMU propagation prior distribution and the posterior distribution resulting from backend pose measurement $\mathbf{T}_{\text{rIFL}} $ in order to achieve maximum a posteriori (MAP) estimation.
The pose $\mathbf{T}_\text{rIFL} = [\mathbf{R}_\text{rIFL}\mid \mathbf{t}_\text{rIFL}]$ consists of a rotation matrix $\mathbf{R}_\text{rIFL} \in SO (3)$ and a translation vector $\mathbf{t}_\text{rIFL} \in \mathbb{R}^3$.
We can directly regard $\mathbf{T}_{\text{rIFL}}$ as an observation of the error state $\delta \mathbf{x} = [\delta \boldsymbol{r}\mid \delta \boldsymbol{t}\mid \delta \boldsymbol{v} \mid \delta \boldsymbol{b_g} \mid \delta \boldsymbol{b_a}\mid \delta \boldsymbol{g}]$. Then the measurement model $\boldsymbol{h}$ is given as follows:
\begin{equation}
\small
\boldsymbol{z}_{\text{rIFL}}=\left[\begin{array}{l}\boldsymbol{z}_{\delta \boldsymbol{r}}\\ \boldsymbol{z}_{\delta \boldsymbol{t}}\end{array}\right] =\left[\begin{array}{l}\boldsymbol{h}(\delta \boldsymbol{r}) \\ \boldsymbol{h}(\delta \boldsymbol{t})\end{array}\right] 
=\left[\begin{array}{c}\log \left({}^G\boldsymbol{R}^{T}_I \boldsymbol{R}_{\mathrm{rIFL}}\right) \\ \boldsymbol{t}_{\text{rIFL}}-{}^G\boldsymbol{t}_I\end{array}\right].
\end{equation}

We have the residual of odometry measurement as 
$r_O= \left[\begin{array}{l}{r}_{\delta\boldsymbol{ r}} \\ {r}_{\delta \boldsymbol{t}}\end{array}\right] \simeq\left[\begin{array}{l}\boldsymbol{H_{O_r}} \delta \widehat{\mathbf{x}}_{i+1}+\boldsymbol{v_r} \\ \boldsymbol{H_{O_t}} \delta \widehat{\mathbf{x}}_{i+1}+\boldsymbol{v_t}\end{array}\right]$, where ${r}_{\delta \theta} $ and ${r}_{\delta t} $
are the rotation and translation errors, $ \boldsymbol{v_r} \in N\left(\mathbf{0}, \boldsymbol{R}_r\right)$ and $\boldsymbol{v_t} \in N\left(\mathbf{0}, \boldsymbol{R}_t\right)$ are the corresponding Gaussian noise. Hence, we can derive the Jacobian matrix for the error state by considering the residuals of rotation and translation:
$$
\begin{aligned}
        \small
\boldsymbol{H_{O_r}} & =\left[\begin{array}{lll}
\boldsymbol{I}_{3 \times 3} & \mathbf{0}_{3 \times 3} & \mathbf{0}_{3 \times 12}
\end{array}\right] \\
\boldsymbol{H_{O_t}} & =\left[\begin{array}{lll}
\mathbf{0}_{3 \times 3} & \boldsymbol{I}_{3 \times 3} & \mathbf{0}_{3 \times 12}
\end{array}\right]
\end{aligned}.
$$

\subsection{Constrained ESIKF State Update}
The CESIKF-based state estimation module uses prior pose and undistorted points to identify the degenerate directions and then calculate additional constraints for robust localization in degenerate scenarios. Finally, RELEAD performs a pose update combining prior LiDAR measurements and additional constraints.

\subsubsection{Scan-Matching Model}
If a LiDAR scan $\mathcal{P}_k$ is received at time $t_k$, RELEAD utilizes a surface measurement model and incorporates the point-to-plane residuals of scan-matching as observation equations:
\begin{equation}
\begin{aligned}
\mathbf{0}=\mathbf{r}_l\left(\check{\mathbf{x}}_{k+1},{ }^L \mathbf{p}_j\right)=\mathbf{u}_j^T\left({ }^G \check{\mathbf{T}}_{I_{k+1}}{ }^I \mathbf{T}_L{ }^L \mathbf{p}_j-\mathbf{q}_j\right).
\end{aligned}
\label{equation:LiDAR_measurement_error}
\end{equation}
With $\check{\mathbf{x}}_{k+1}$ being the current estimate of $\mathbf{x}_{k+1}$, we can transform ${ }^L \mathbf{p}_j \in \mathcal{P}_k $ from LiDAR frame to the global frame ${ }^G \mathbf{p}_j$. We search for the nearest points in the map and use them to fit a plane with normal $\mathbf{u}_j$ and an in-plane point $\mathbf{q}_j$.
When LiDAR degeneration occurs, the associated states become unrestricted and vulnerable to measurement noise. To tackle this issue, we first identify the deteriorated directions before the optimization procedure and introduce additional constraints along these directions.

\subsubsection{Degeneracy Detection}
\label{subsubsection:Degeneracy_Detection}
\begin{figure}[t]
        \centering
        \subfloat[ A corridor example illustrating the degeneration detection. Points $\boldsymbol{p}$ (green arrows), surface normals $\boldsymbol{n}$ (red arrows), the robot center (black dot).]{\includegraphics[width=0.22\textwidth]{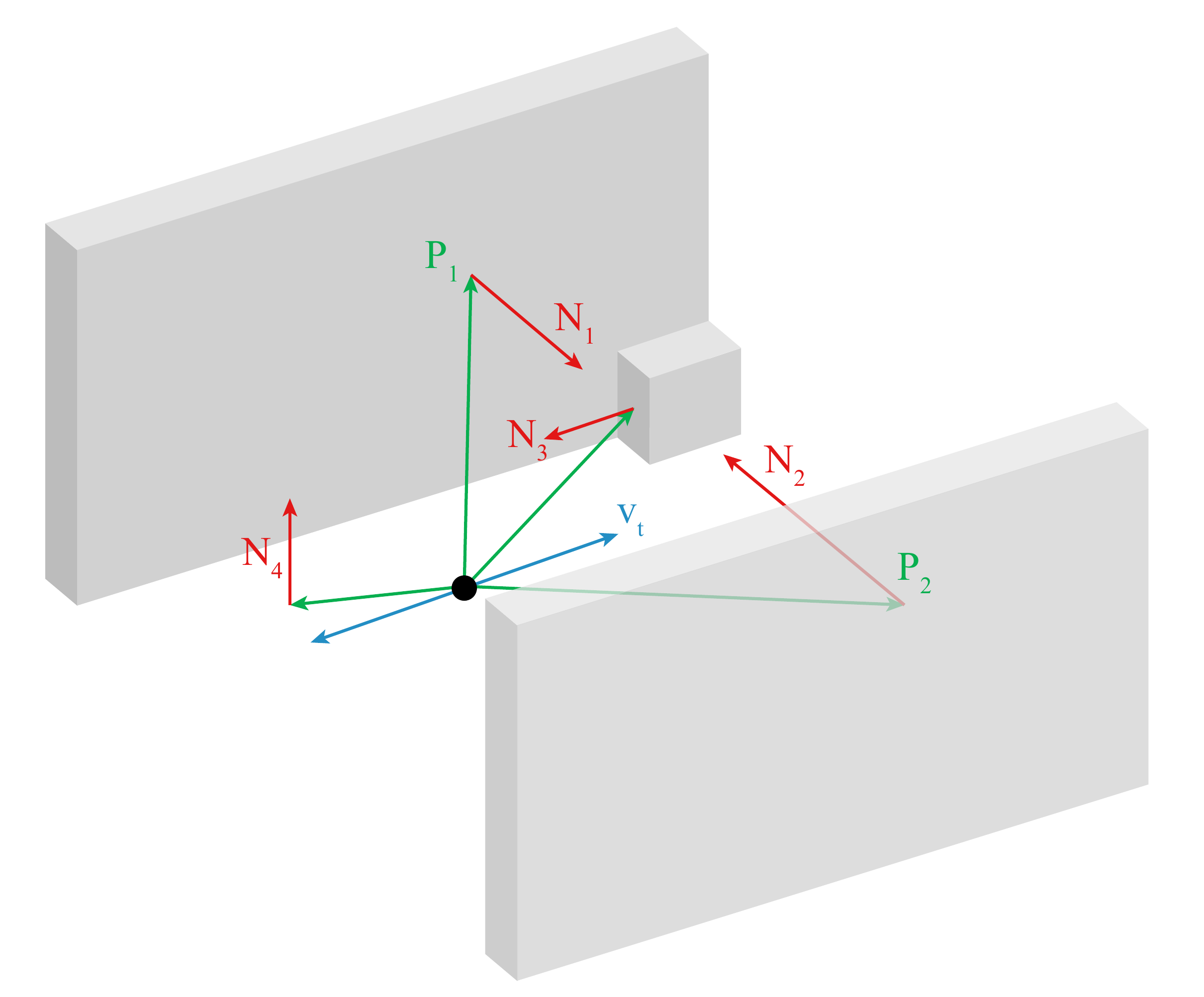}
        \label{fig:Degeneration-Dectection}}
        \hfil
        \subfloat[The increment of point cloud alignment from the initial value is projected onto the constraint plane using CESIKF.]{\includegraphics[width=0.22\textwidth]{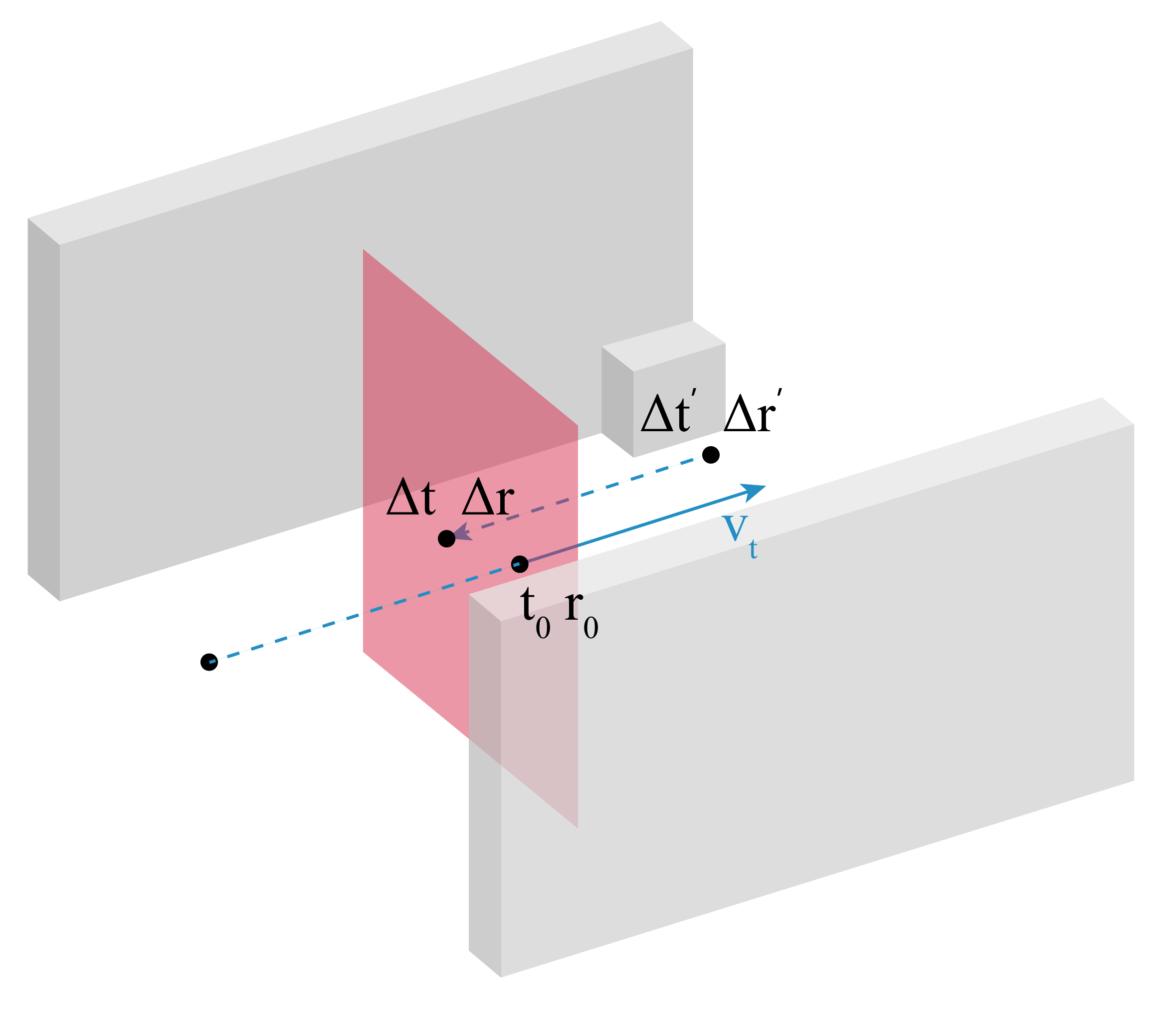}
        \label{fig:Degeneration-Aware}}
        \vspace{-0.2cm} 
        \caption{Degeneration-Aware LiDAR Measurement Update.}
        \label{fig:Degeneration_Detection}
        \vspace{-0.6cm} 
\end{figure}

Utilizing the prior, undistorted points and the global map, we can conduct degeneration detection as illustrated in Fig.\ref{fig:Degeneration_Detection}(a) by assessing the strength of the information pairs $(_{L}\boldsymbol{p}, _{L}\boldsymbol{u})$. We can reformulate the error function (\ref{equation:LiDAR_measurement_error}) as a quadratic cost optimization problem, so the Hessian matrix $\boldsymbol{A}^{\prime}$ can be divided as follows:
\begin{equation}
        \footnotesize
\boldsymbol{A}^{\prime}= \underbrace{(\sum_{i=1}^N \underbrace{\left[\begin{array}{c}
        \left(\boldsymbol{p}_i \times \boldsymbol{n}_i\right) \\
        \boldsymbol{u}_i
        \end{array}\right]}_{\boldsymbol{A}} \underbrace{\left[\left(\boldsymbol{p}_i \times \boldsymbol{u}_i\right)^T \boldsymbol{n}_i^T\right]}_{\boldsymbol{A}^{\top}})}_{\boldsymbol{A}^{\prime}}
         =\left[\begin{array}{ll}
\boldsymbol{A}_{r r}^{\prime} & \boldsymbol{A}_{r t}^{\prime} \\
\boldsymbol{A}_{t r}^{\prime} & \boldsymbol{A}_{t t}^{\prime}
\end{array}\right]_{6 \times 6}.
\end{equation}

And then we focus on the eigenanalysis of $\boldsymbol{A}_{t t}^{\prime}$ and $\boldsymbol{A}_{r r}^{\prime}$ by using SVD. For the rotation and translation components, the eigendecomposition is given as
$
\boldsymbol{A}_{t t}^{\prime}=\boldsymbol{V}_t \Sigma_t \boldsymbol{V}_t^{\top}, \quad \boldsymbol{A}_{r r}^{\prime}=\boldsymbol{V}_r \Sigma_r \mathbf{V}_r^{\top}.
$
By utilizing the X-ICP method \cite{Tuna2022XICPLL}, RELEAD performs an analysis on the impacts of constraints along the translation vector $\mathbf{V}_t$ and rotation vector $\mathbf{V}_r$. This analysis employs information pairs to determine the localizability vector, which indicates the directions that exhibit degeneracy.
$
\Omega =\left(\left\{\Omega_{{}_L v_{t_1}}, \Omega_{{}_L v_{t_2}}, \Omega_{{}_L v_{t_3}}\right\},\left\{\Omega_{{}_L v_{r_1}}, \Omega_{{}_L v_{r_2}}, \Omega_{{}_L v_{t_3}}\right\}\right)^{\top}
$, 
where $\left\{{ }_L v_{t_1}, {}_L v_{t_2}, {}_L v_{t_3}\right\} \in V_{\mathrm{t}}$, $\left\{{ }_{L} \boldsymbol{v}_{r_1},{ }_{L} \boldsymbol{v}_{r_2},{ }_{L} \boldsymbol{v}_{r_3}\right\} \in$ $V_r$ and $\Omega_i \in$ \{none, partial, full\}. So, we now get the categories corresponding to every direction being nonlocalizable, partially-localizable, and localizable, respectively.

\subsubsection{Constrained Optimization}
To incorporate the degeneration information from Section \ref{subsubsection:Degeneracy_Detection} consistently, we transform the translation $\mathbf{V}_t$ and rotation $\mathbf{V}_r$ directions with prior pose to the world frame. Subsequently, we incorporate additional constraints into the ESIKF update using a methodology akin to the optimization module in X-ICP:
\begin{equation}
        \small
\begin{aligned}
\boldsymbol{v}_j^{\top} \cdot\left(\delta \boldsymbol{t}-\boldsymbol{t}_0\right)=0, & \text { if } \boldsymbol{v}_j \in \boldsymbol{V}_t, \\
\boldsymbol{v}_j^{\top} \cdot\left(\delta \boldsymbol{r}-\boldsymbol{r}_0\right)=0, & \text { if } \boldsymbol{v}_j \in \boldsymbol{V}_r .
\end{aligned}
\label{equation:Constrained_Optimization}
\end{equation}

The constraint equation aims to resample the most constrained part of the point cloud in the degradation direction for alignment. As shown in Fig. \ref{fig:Degeneration_Detection} (b), an update $\delta t, \delta r $ satisfying these constraints is derived by projecting the EISKF-solved pose update $\delta t', \delta r'$ onto the constraint plane. 
This projection considers equality constraints $\mathbf{C} \Delta \mathbf{x}= \mathbf{d}$ from Equation. (\ref{equation:Constrained_Optimization}), which can be represented as:
\begin{equation}
        \small
        \underbrace{\left[\begin{array}{cc}
        \boldsymbol{v}_j  & \mathbf{0}_{m_r \times 3}\\
        \vdots & \vdots \\
        \mathbf{0}_{m_t \times 3}  & \boldsymbol{v}_j
        \end{array}\right]}_{\mathbf{C}_{\left(m_r+m_t\right) \times 6}} 
        \underbrace{\left[\begin{array}{c}
               \delta  \boldsymbol{r}\\
               \delta  \boldsymbol{t}
        \end{array}\right]}_{\Delta \mathbf{x}}
        =\underbrace{\left[\begin{array}{c}
        \boldsymbol{v}_j \cdot \boldsymbol{r}_0 \\
        \vdots \\
        \boldsymbol{v}_j \cdot \boldsymbol{t}_0
        \end{array}\right]}_{\mathbf{d}_{\left(m_r+m_t\right) \times 1}},
\end{equation}
where $m_r$ and $m_t$ represent the count of degenerate directions in the rotation and translation spaces, respectively.
By employing the Lagrange multiplier method following the LiDAR measurement update, a constrained update for $\Delta \mathbf{x}$ component of $\delta \mathbf{x}$ can be implemented effectively:
\begin{equation}
        \small
        \begin{aligned}
                \mathbf{G} &= \mathbf{\Sigma} \mathbf{C}^T(\mathbf{C} \mathbf{\Sigma} \mathbf{C}^T)^{-1}, \\
        \Delta\overline{\mathbf{x}} &= \Delta\hat{\mathbf{x}} - \mathbf{G} (\mathbf{C} \Delta \hat{\mathbf{x}} - \mathbf{d}). \\
        \end{aligned}
\end{equation}

\subsection{Flexible Graph-based Backend}
\label{subsection:backend}
\subsubsection{Delay-tolerant Architecture}
RELEAD's backend uses a conditionally independent pipeline, connecting odometry measurements via IMU preintegration factors. This IMU-centric sensor fusion approach allows RELEAD to handle transient data loss and integrate various sensor types effectively. As long as not all modalities fail simultaneously, IMU bias can be constrained by any relative pose, ensuring continuous and accurate state estimation. However, this pipeline must address challenges related to unsynchronized odometry sources due to varying processing delays.

To circumvent this issue, RELEAD establishes a new state node in the pose graph for each IMU measurement, allowing every incoming relative odometry measurement at a lower rate to link to a specific node created at a higher rate with minimal time difference. In this configuration, RELEAD fundamentally eliminates delayed measurement processing. Specifically, by updating the state precisely at the sampling time of each IMU, the graph avoids the problem of lacking anchor nodes for earlier measurement updates. 

For each new odometry measurement, we minimize an optimization problem that consists of relative pose factor $\mathbf{e}_{i j}^{odom}$ and preintegrated IMU factors $\mathbf{e}_{i j}^{imu}$ as follows:
\begin{equation}
        \small
\begin{aligned}
\chi^* = \mathop{\arg\min}\limits_{\chi} \sum_{i, j} \left( \|\mathbf{e}_{i j}^{odom}\|_{W_{i j}}^2  + \|\mathbf{e}_{i j}^{imu}\|_{W_{i j}}^2 \right),
\end{aligned}
\end{equation}
where $i$ and $j$ serve as indexes spanning all factors.

\begin{figure*}[t]
        \centering
        \subfloat[Playground\_2 $\alpha$]{\includegraphics[width=0.33\textwidth]{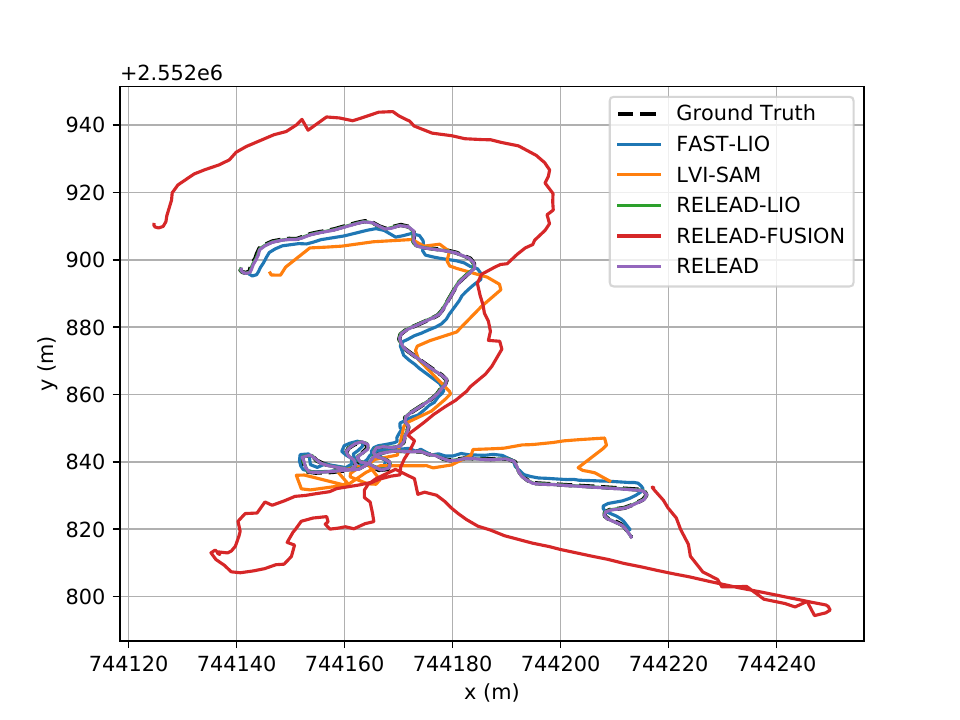}
        \label{fig:S3EAlpha}}
        \subfloat[Playground\_2 $\beta $]{\includegraphics[width=0.33\textwidth]{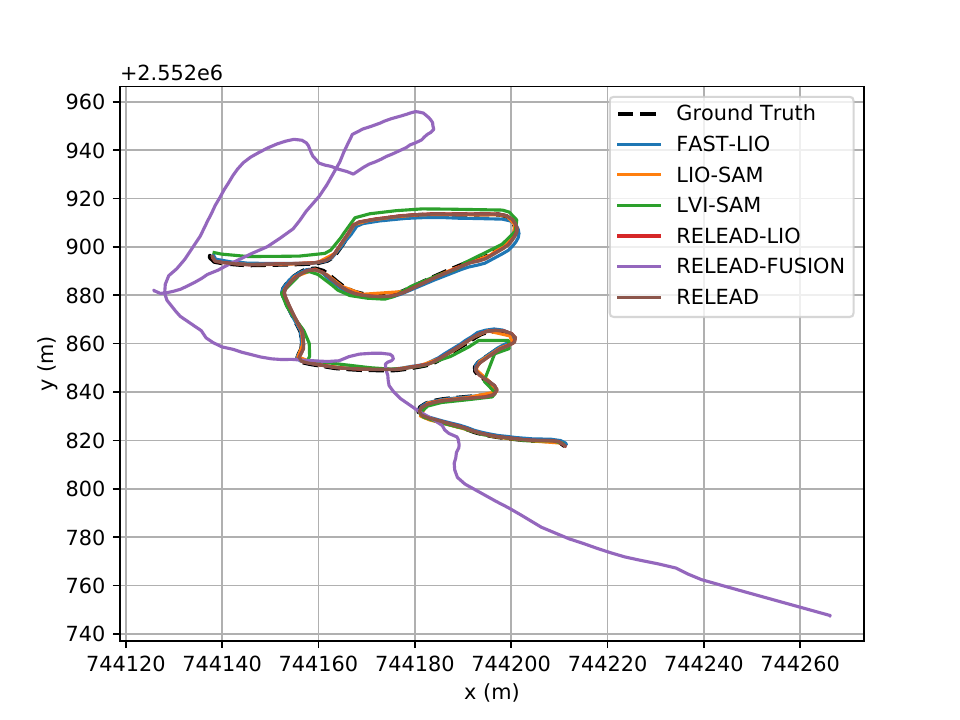}
        \label{fig:S3EBob}}
        \subfloat[Playground\_2 $\gamma  $]{\includegraphics[width=0.325\textwidth]{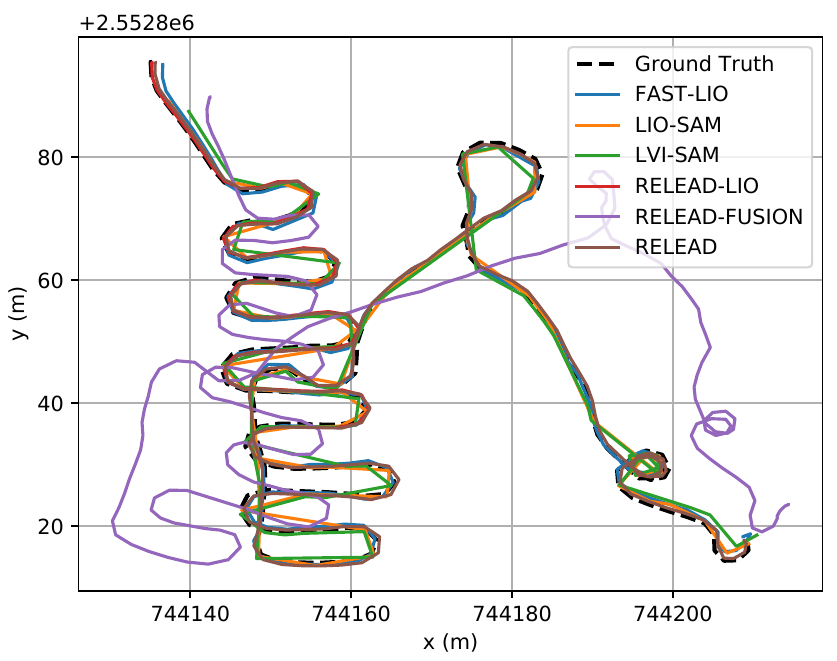}
        \label{fig:S3ECarol}}
        \caption{Trajectory results for three sequences in \emph{S3E} dataset.}
        \label{fig:S3E}
        \vspace{-0.4cm} 
\end{figure*}

\subsubsection{GNC-based Outlier Rejection Strategy} The other problem of the conditionally independent pipeline is that the sensor fusion architecture is brittle to outliers due to sensor degradation or environmental changes. RELEAD employs a graduated non-convexity factor methodology to counteract the influence of outlier odometry measurements effectively. We consider the IMU factor and LiDAR odometry measurements without degeneration as known inliers, while odometry measurements from other modalities and LiDAR odometry with degeneration are potential outliers. 
According to \cite{yang2020graduated}, GNC is a method developed to address the challenges of optimizing the robust cost function. By incorporating GNC, we can improve the initial stage of solving non-convex variations of the problem, thereby reducing the likelihood of converging towards a local optimum. So, we can formulate the robustified objective function as follows:
\begin{equation}
        \small
\label{equation:GNC}
\begin{aligned}
\chi^* = \mathop{\arg\min}\limits_{\chi} \sum_{i, j} \left( \|\mathbf{e}_{i j}^{imu}\|_{W_{i j}}^2  + \varPsi \left(\|\mathbf{e}_{i j}^{odom}\|_{W_{i j}}^2 \right) \right)
\end{aligned}
\end{equation}
where the function $\varPsi (*)$ is a robust function, here we use the scale-invariant graduated (SIG) kernel function \cite{mcgann2023robust} for online efficiency.

\subsection{Robust Incremental Fixed Lag Smoother}
Two primary challenges hinder the attainment of precise solutions to Eq.(\ref{equation:GNC}) when dealing with outlier measurements while maintaining online efficiency: i) GNC solving typically involves a batch process and iterative optimization routines, rendering it inefficient and unsuitable for RELEAD. ii) RELEAD's method of creating state nodes results in a rapidly growing graph, which can eventually become too large for online optimization.

To tackle the former problem, we adopted the incremental graduated optimization method riSAM \cite{mcgann2023robust} to acquire an efficient form of Graduated Non-Convexity. 
Although riSAM improves the feasibility of solving the GNC optimization, it still cannot perform real-time computation for the large-scale pose graph in RELEAD. Our solution involves using marginalization to constrain the problem size. We incorporate the marginalization technique proposed in \cite{chiu2013robust} with riSAM to retain only a single pose within a sliding window of 1 second while marginalizing the remaining poses. For simplicity of marginalization, we use the duplicate elimination ordering as \cite{kaess2012isam2} instead of the custom ordering for better runtime efficiency in riSAM. 
This gives rise to the sliding-window invariant riSAM, enabling the transformation of the objective function in Eq.(\ref{equation:GNC}) into the subsequent expression:
\begin{equation}
        \small
\label{equation:rIFL}
\begin{aligned}
\chi^* = \mathop{\arg\min}\limits_{\chi} \sum_{i, j} \left( \|\mathbf{e}_{i j}^{imu}\|_{W_{i j}}^2  + \varPsi \left(\|\mathbf{e}_{i j}^{odom}\|_{W_{i j}}^2 \right) \right) + \mathbf{E}_{\mathrm{m}}
\end{aligned}
\end{equation}
where $\mathbf{E}_m$ represents the marginalization prior from the robust Incremental Fixed Lag Smoother (rIFL). 
                                                                        
%

\section{EXPERIMENTS AND RESULTS}
In this section, we perform two experiments to demonstrate the robustness of RELEAD by utilizing the S3E dataset \cite{Feng2022S3EAL} and self-collected tunnel data.
The first experiment highlights RELEAD's ability to handle outlier measurements, while the second demonstrates its accurate localization in degenerate scenarios. We evaluate our method in comparison to the most advanced LiDAR-inertial odometry methods: LIO-SAM \cite{Shan2020LIOSAMTL}, FAST-LIO2 \cite{Xu2021FASTLIO2FD}, and LiDAR-visual-inertial odometry methods: LVI-SAM \cite{Shan2021LVISAMTL} and R3LIVE \cite{lin2022r3}.

\begin{table}[t]
        \small
        \caption{ABSOLUTE TRANSLATION ERROR (RMSE, METERS) ON \emph{S3E} DATASET}
        \label{table:OutlierMeasurement}
        \scriptsize
        \centering
        \begin{threeparttable} 
        \begin{tabular}{c
        >{\columncolor[HTML]{EFEFEF}}c c
        >{\columncolor[HTML]{EFEFEF}}c }
        \toprule
        Methods       & Playground\_2 $\alpha $ & Playground\_2 $\beta $ & Playground\_2 $\gamma $ \\ \midrule
        LIO-SAM       & $\times$      & $\boldsymbol{0.36}$         & 0.60          \\
        LVI-SAM       & 6.78          & 6.10          & 0.72          \\
        FAST-LIO      & 1.92          & 0.94          & 0.73          \\
        R3LIVE        & $\times$      & $\times$      & $\times$      \\ \midrule
        RELEAD-LIO\tnote{*}    & $\boldsymbol{0.12}$          & $\boldsymbol{0.36}$          & $\boldsymbol{0.37}$          \\
        RELEAD-Fusion\tnote{*} & 15.16         & 15.13         & 9.49          \\
        RELEAD\tnote{*}        & 0.31          & 0.49          & 0.66          \\ \bottomrule
        \end{tabular}
        \begin{tablenotes} 
        \item[*] RELEAD represents the full proposed method, RELEAD-LIO means the LIO part of RELEAD, and RELEAD-fusion means the multisensor fusion system without using GNC-factor. $\times$ means that the SLAM method failed. 
        \end{tablenotes}
        \end{threeparttable}
\end{table}

\subsection{Outlier Measurement}
The first experiment evaluates RELEAD's robustness in handling outlier measurements using S3E dataset. These sequences involve ground vehicles performing substantial maneuvers in an empty playground. The challenging aspect of these scenarios arises from the significant impact of camera exposure changes on VIO's pose estimation, resulting in outlier measurements for sensor fusion. Results for ATE and trajectories are presented in Tab.\ref{table:OutlierMeasurement} and Fig.\ref{fig:S3E}, respectively.

LIO-SAM exhibits the slightest error in the $\beta$ sequence and the second slightest error in the $\gamma$ sequence. 
However, LIO-SAM fails to obtain accurate estimations and gets dispersion in the $\alpha$ sequence.
The fusion of visual information allows LVI-SAM to maintain continuous pose estimation in the $\alpha$ sequences compared to LIO-SAM.  However, LVI-SAM's visual outlier poses as initial values result in less favorable outcomes than LIO-SAM in both the $\beta$ and $\gamma$ sequences.
FAST-LIO exhibits superior robustness compared to LIO-SAM and LVI-SAM, resulting in more minor errors than LVI-SAM across all three sequences. 
However, R3LIVE's tightly coupled architecture struggled to correct inaccurate camera pose estimates, leading to even lower resilience than FAST-LIO. 
Consequently, R3LIVE failed to achieve successful localization in all three sequences.

\begin{figure*}[t]
        \centering
        \subfloat[Mapping results: \uppercase\expandafter{\romannumeral1}) LVI-SAM, \uppercase\expandafter{\romannumeral2}) R3LIVE, \uppercase\expandafter{\romannumeral3}) RELEAD, \uppercase\expandafter{\romannumeral4}) LIO-SAM, \uppercase\expandafter{\romannumeral5}) FAST-LIO.]{\includegraphics[width=0.7\textwidth]{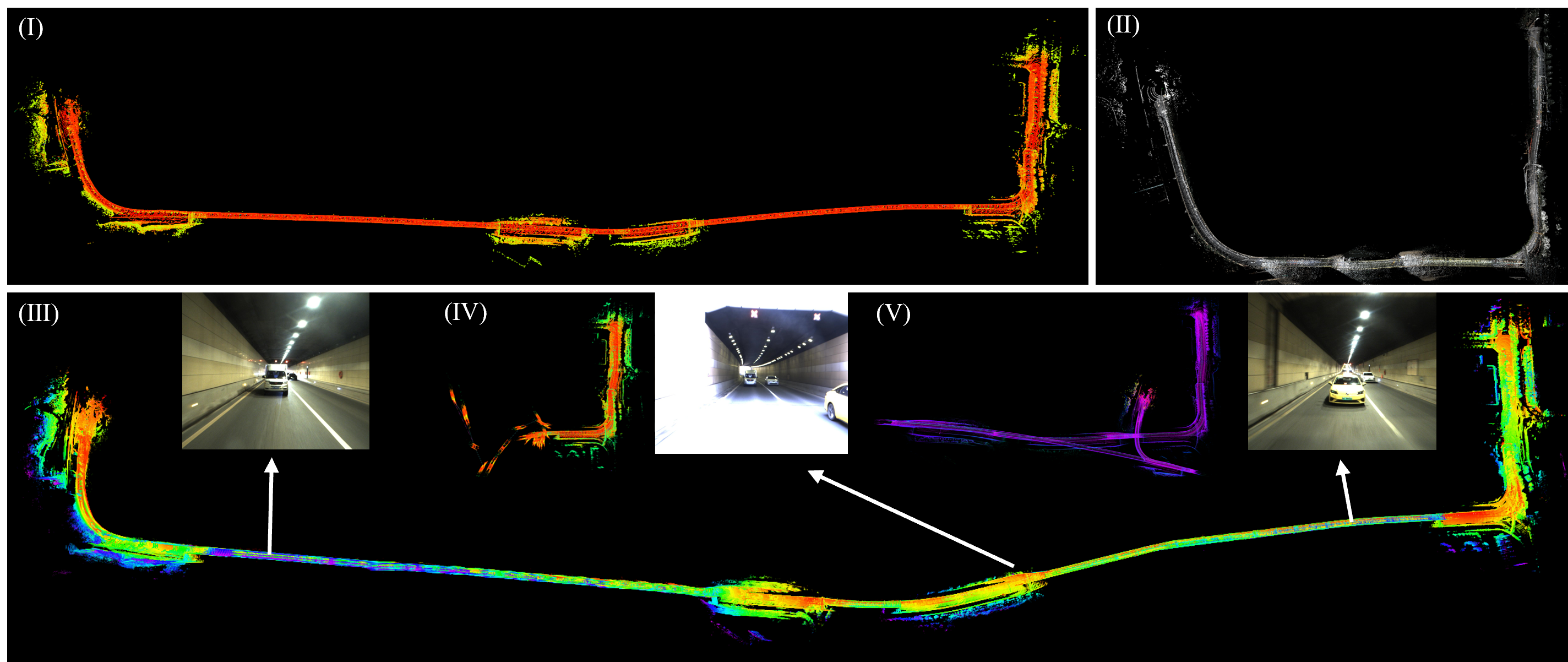}
        \label{fig:tunnel01-mapping}}
        \subfloat[Trajectory results]{\includegraphics[width=0.23\textwidth]{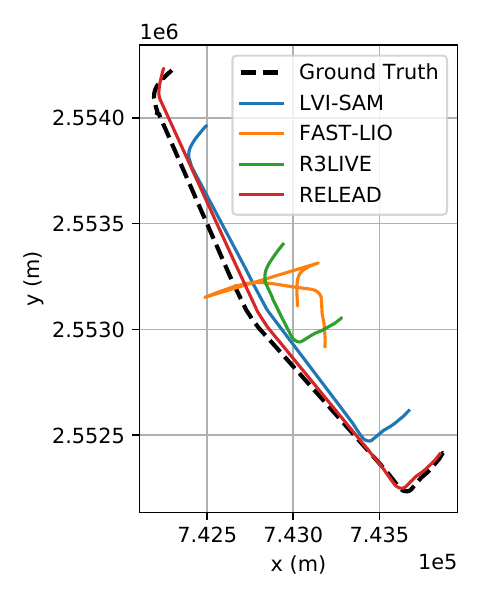}
        \label{fig:tunnel01-trajectories}}
        \caption{Mapping and trajectory results of experiment in urban road tunnel.}
        \label{fig:tunnel01}
        \vspace{-0.5cm} 
\end{figure*}

RELEAD-LIO, featuring a continuous-time point cloud de-distortion model, as well as degradation detection and constraint optimization modules, effectively addresses significant maneuvers and minor degradation.
RELEAD-FUSION integrates VIO pose data from VINS-MONO; however, an indiscriminate fusion of outliers and inliers would yield unfavorable outcomes.
With the aid of the GNC factor, RELEAD excels in removing inaccurate VIO poses during backend fusion, achieving results on par with LIO. However, it relies on a fixed uncertainty approach for multi-sensor fusion, which does not adapt to changing environments or evolving motion patterns. As a result, in specific scenarios, RELEAD's outcomes might exhibit slightly reduced accuracy compared to LIO.

\subsection{Geometrically Self Similar Environment}
\subsubsection{Scenario 1: Metro Tunnels}
In this scenario, the ground vehicle continuously follows a slightly right-curving circular tunnel. Mapping results are shown in Fig.\ref{fig:shenzhentunnel}. Among the tested methods, RELEAD consistently produces a forward trajectory, while others yield reverse pose estimates. FAST-LIO faces localization challenges in this scenario. Although LIO-SAM utilizes a solution remapping approach from \cite{Zhang2016OnDO} to address degeneracy issues, it still fails to obtain accurate pose estimations. 
Regarding LVI-SAM, the vision sub-module depends on LIO's output for initialization, and inaccuracies in LIO estimation adversely affect the performance of the vision module. Moreover, LVI-SAM utilizes the vision's estimation as the initial value for LIO, creating a nested approach that exacerbates system failures in degraded tunnel scenarios. The subway tunnels' highly uniform textures and colors pose a significant challenge for R3LIVE's visual photometric error optimization. Consequently, R3LIVE's VIO module fails to effectively assist LIO in resolving the degradation issue.

RELEAD excels in this test. Its success is attributed to incorporating pose information from the backend, fusing visual data to improve IMU propagation outcomes, and thus, supplying robust initial values for LIO updates. Additionally, including degradation detection and constraint optimization modules ensures minimal noise in point cloud registration along the degradation direction.

\begin{figure}[t]
        \centering
        \vspace{-0.1cm} 
        \includegraphics[width=0.45\textwidth, height=0.185\textwidth]{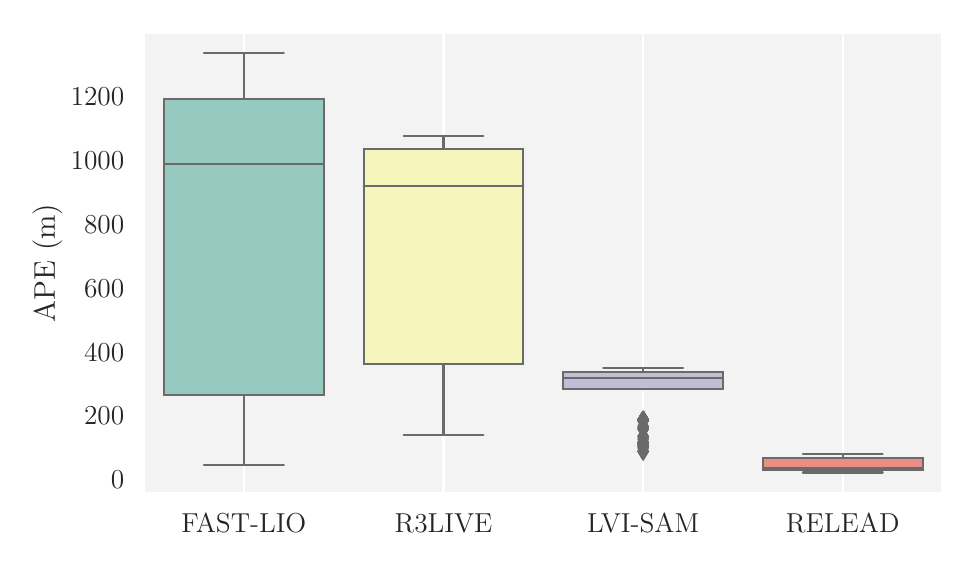}
        \caption{ATE on urban road tunnels dataset}
        \label{fig:ATEontunnel01}
        \vspace{-0.7cm} 
\end{figure}

\subsubsection{Scenario 2: Urban Road Tunnels}
To assess the effectiveness of our proposed method in extensive settings, we performed evaluations within urban road tunnels. This environment comprises extensive, highly degraded tunnels covering a total distance of 2942 meters. The changing camera exposure as vehicles enter and exit the tunnel poses challenges for visual localization. Results for ATE, trajectories, and mapping are presented in Fig.\ref{fig:tunnel01} and Fig.\ref{fig:ATEontunnel01}, respectively.
Similar to the subway tunnel experiment, LIO-SAM, and FAST-LIO failed to establish accurate positioning due to geometrically self-similar structures.
Even with the inclusion of visual data, LVI-SAM managed to maintain a continuous output of pose estimations compared to LIO-SAM. However, it still struggled to provide precise positioning results within the tunnels. The results from R3LIVE mirrored those of LVI-SAM. While introducing visual information prevented it from producing backward pose estimations like FAST-LIO, the system still encountered difficulties in this tunnel scenario.

In stark contrast, RELEAD outperformed the other methods by achieving minor errors and delivering commendable mapping results. Incorporating GNC factors guarantees removing outliers from the backend, facilitating the attainment of accurate estimates in degenerated scenarios.
Meanwhile, RELEAD's backend can seamlessly integrate various sensor information. Incorporating GNSS data enhanced positioning accuracy, resulting in an error of only 0.5 meters.

\begin{figure}[t]
        \centering
        \vspace{-0.1cm} 
        \includegraphics[width=0.45\textwidth,]{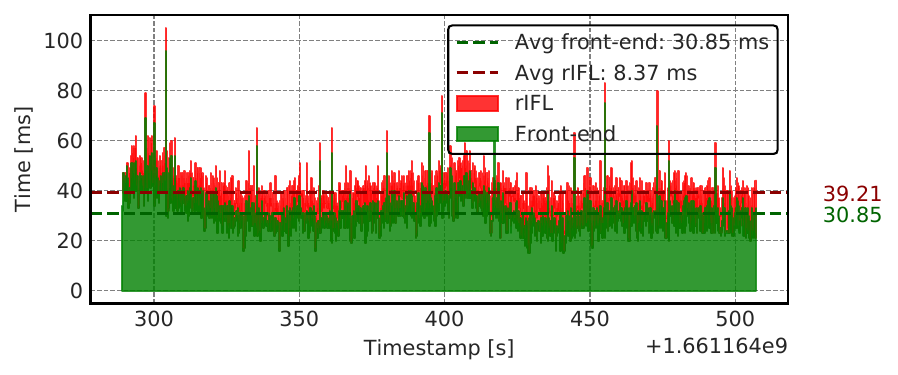}
        \vspace{-0.2cm} 
        \caption*{\small Specification: Intel i7 CPU@4.90 GHz and 16GB RAM}
        \vspace{-0.2cm} 
        \caption{Average execution time on sequence Playground\_2$\alpha $}
        \label{fig:algorithm_time_analysis1}
        \vspace{-0.6cm} 
\end{figure}

\subsection{Time Analysis}
We employ front-end and back-end execution times per scan of incremental updates as performance metrics to assess the efficiency of RELEAD. As depicted in Fig.\ref{fig:algorithm_time_analysis1}, the outcomes demonstrate that RELEAD exhibits exceptional real-time capabilities (Avg. = 39.21 ms). We also computed the average runtime without the GNC factor and with an expanded sliding window to 2s, yielding 28.65ms and 44.65ms, respectively. Our rIFL solver effectively mitigates the computational burden of generating state nodes for each IMU measurement and employing GNC factors.

                                                                        
%
\section{CONCLUSION}
This study presents RELEAD, a multisensor-enhanced LiDAR odometry that facilitates dependable and robust pose estimation in diverse environments with LiDAR degradation. The proposed approach excels in identifying environmental degradation and achieving precise, noise-free pose estimation in the degradation direction through constrained optimization. Additionally, by incorporating GNC factors, it effectively integrates multiple sensors while robustly handling outlier measurements. The resilience of RELEAD is demonstrated through various experiments.
Currently, our method employs time-constant error models with fixed covariance during the fusion of multisensory data in the backend. In the future, we intend to implement real-time estimation of measurement covariance, similar to \cite{fakoorian2022rose}, further enhancing accuracy in multisensor fusion and eliminating outliers.



\clearpage
\bibliographystyle{IEEEtran}
\bibliography{root}

\end{document}